\documentclass[conference]{IEEEtran}
\IEEEoverridecommandlockouts
\usepackage{amsmath,amssymb,amsfonts}
\usepackage{algorithmic}
\usepackage{graphicx}
\usepackage{textcomp}
\usepackage{hyperref}
\usepackage{booktabs}
\usepackage[table]{xcolor}
\usepackage{graphicx}
\usepackage[backend=biber]{biblatex}
\usepackage{booktabs}
\usepackage{longtable}
\usepackage{array}
\usepackage{ragged2e}

\usepackage{hyperref}

\def\BibTeX{{\rm B\kern-.05em{\sc i\kern-.025em b}\kern-.08em
    T\kern-.1667em\lower.7ex\hbox{E}\kern-.125emX}}
\begin{document}

\title{Comparative Analysis of Stroke Prediction Models Using Machine Learning\\}

\author{
\centering
\IEEEauthorblockN{
    \textsuperscript{1} Anastasija Tashkova, 
    \textsuperscript{2} Stefan Eftimov, 
    \textsuperscript{3} Bojan Ristov, 
    \textsuperscript{4} Slobodan Kalajdziski, PhD
} 
\\
\IEEEauthorblockA{
    \textit{Faculty of Computer Science and Engineering}, 
    \textit{Ss. Cyril and Methodius University} \\
    Skopje, North Macedonia
} 
\\
\IEEEauthorblockA{
    \textsuperscript{1} anastasija.tashkova@students.finki.ukim.mk, 
    \textsuperscript{2} stefan.eftimov.1@students.finki.ukim.mk, \\
    \textsuperscript{3} bojan.ristov@students.finki.ukim.mk, 
    \textsuperscript{4} slobodan.kalajdziski@finki.ukim.mk
    }
}


\maketitle

\begin{abstract}

Stroke remains one of the most critical global health challenges, ranking as the second leading cause of death and the third leading cause of disability worldwide. This study explores the effectiveness of machine learning algorithms in predicting stroke risk using demographic, clinical, and lifestyle data from the Stroke Prediction Dataset. By addressing key methodological challenges such as class imbalance and missing data, we evaluated the performance of multiple models, including Logistic Regression, Random Forest, and XGBoost. Our results demonstrate that while these models achieve high accuracy, sensitivity remains a limiting factor for real-world clinical applications. In addition, we identify the most influential predictive features and propose strategies to improve machine learning-based stroke prediction. These findings contribute to the development of more reliable and interpretable models for the early assessment of stroke risk.
\end{abstract}


\section{Introduction}

\subsection{Global Burden of Stroke}

Stroke is a neurological condition caused by a disruption of blood supply to the brain, leading to rapid cell death and potential long-term disability. The Global Burden of Disease study reports over 7.25 million stroke-related deaths in 2021, with ischemic stroke contributing the largest share. Stroke affects 12 million people annually, and the lifetime risk is one in four individuals over the age of 25. This condition also incurs substantial economic costs, estimated at over US \$890 billion, representing 0.66\% of the global GDP \cite{1}.

\subsection{Machine Learning in Healthcare}

The growth of electronic health records (EHRs) has opened new possibilities for machine learning in healthcare, enabling the identification of complex patterns in large datasets. Machine learning is increasingly used in diagnostic support, risk assessment, and patient monitoring \cite{2}. However, healthcare data poses challenges such as class imbalance, missing data, and the need for interpretable models.

\subsection{Research Gap and Objectives}

While machine learning has been applied to stroke prediction, comparisons of multiple algorithms in this context remain limited. Previous studies, including (PMC11106277, 2021) and (Tugas Kelompok BI, 2021) \cite{3} \cite{4}, have explored predictive models but often do not comprehensively address class imbalance, potentially leading to overestimated accuracy. Additionally, these studies lack a detailed feature importance analysis to determine the most influential factors in stroke prediction. To support this analysis, we used a publicly available dataset of 5,110 patient records, covering demographic, clinical, and lifestyle variables.

Building on these works, our study extends prior research by systematically evaluating multiple machine learning models (Logistic Regression, Random Forest, Decision Tree, SVM, XGBoost) while implementing and comparing oversampling, undersampling, and SMOTE techniques to mitigate class imbalance. Furthermore, we conduct an in-depth feature importance analysis to enhance interpretability and improve the reliability of stroke risk assessment models.

\section{Data Preprocessing Pipeline}

\subsection{Dataset Description}

The Stroke Prediction Dataset \cite{5} consists of 5,110 patient records with 12 attributes related to demographic, lifestyle, and clinical factors. The target variable is binary, indicating whether a patient has experienced a stroke (1) or not (0). The dataset presents a significant class imbalance, with only 249 positive cases (4.87\%) compared to 4,861 negative cases (95.13\%).

\subsection{Feature Categories}

The dataset includes the following features:

\begin{itemize}
    \item Demographic features: Age, gender, marriage status
    \item Lifestyle factors: Smoking status, work type, residence type
    \item Clinical indicators: Hypertension, heart disease, average glucose level, BMI
\end{itemize}

\subsection{Handling Missing Values}
Initial data exploration revealed 201 missing values in the BMI column (3.93\% of the dataset). Rather than removing these records, which would reduce the already limited number of positive cases, an Iterative Imputer was applied using a Random Forest Regressor \cite{6}. This approach predicts missing BMI values based on patterns in the dataset, leveraging relationships with other features rather than relying on a simple statistical measure like the mean or median. 

\subsection{Categorical Variable Encoding}\label{AA}
Categorical variables were transformed into numerical representations to make them compatible with machine learning algorithms. The encoding strategy varied based on the nature of each variable:
\begin{itemize}
    \item Binary encoding was applied to gender, hypertension, heart disease, ever married, residence type and stroke.
    \item Ordinal encoding was implemented for smoking status (Unknown=0, Formerly smoked=1, Never smoked=2, Smokes=3) based on potential risk level.
    \item Label encoding was utilized for work type, which has five categorical values (Private, Self-employed, Govt\_job, children, Never\_worked), to avoid imposing artificial ordinality.
\end{itemize}

\subsection{Addressing Class Imbalance}
The severe class imbalance (4.87\% stroke cases) presents a significant challenge for model training, potentially leading to high accuracy but poor sensitivity. To address this issue, a combination of \textbf{Oversampling}, \textbf{Undersampling} and \textbf{SMOTE} was applied \cite{7} \cite{8}.

Oversampling was used to increase the number of stroke cases by randomly duplicating existing minority class samples. SMOTE (Synthetic Minority Oversampling Technique) further improved minority class representation by generating synthetic examples rather than just duplicating existing ones. Undersampling was applied to reduce the dominance of the majority class by randomly removing some non-stroke cases.

By combining these techniques, the dataset was balanced more effectively, improving the model’s ability to detect stroke cases without introducing bias towards the majority class.

\section{Analysis of Key Variables}

\begin{figure}[t]
\centering
\includegraphics[width=\columnwidth]{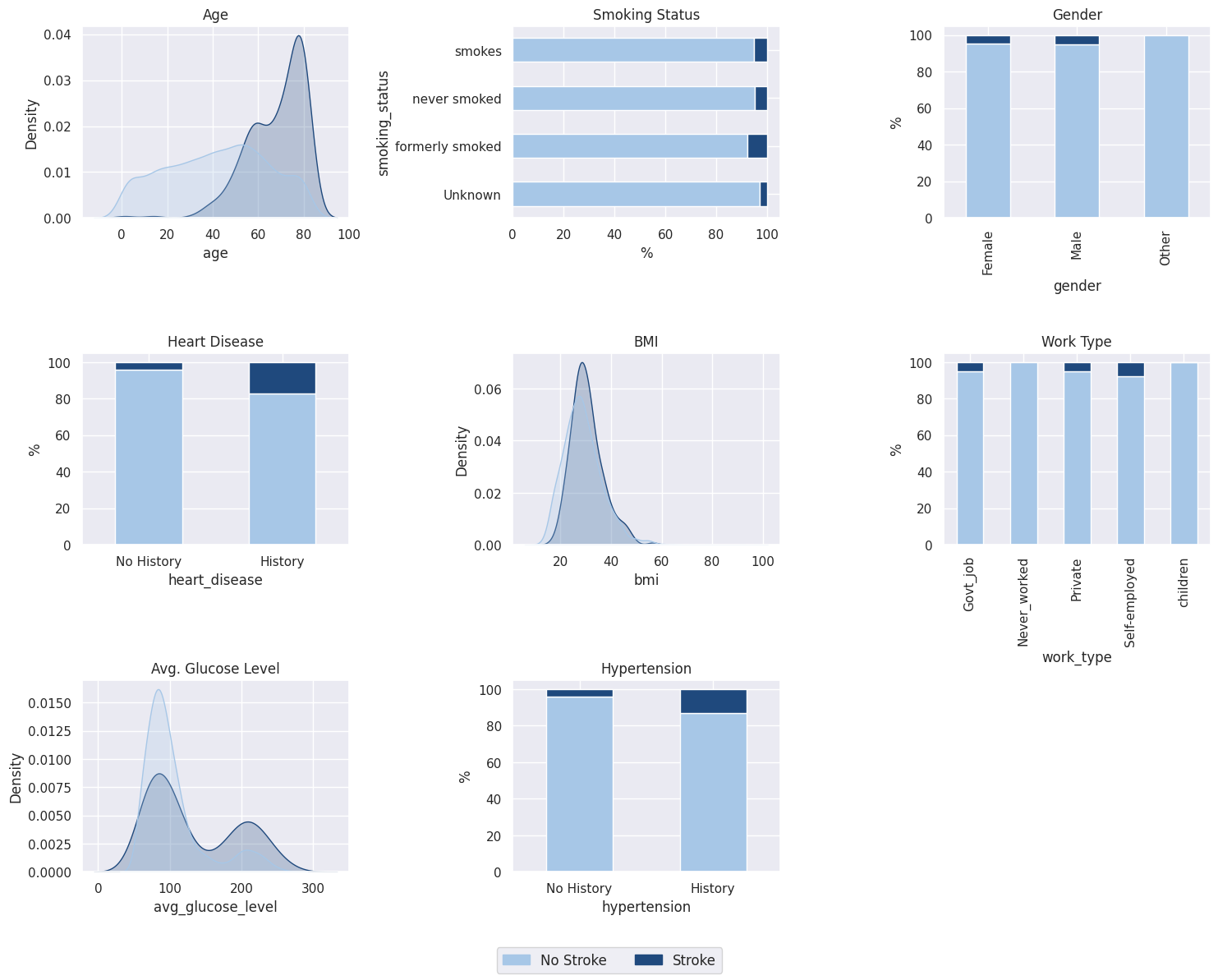}
\caption{Exploratory Data Analysis (EDA) of Stroke Prediction Dataset}
\label{fig:EDA}
\end{figure}

The exploratory data analysis (EDA) examines key variables in the stroke prediction dataset, revealing significant patterns relevant to predictive modeling (Fig. \ref{fig:EDA}). 

Stroke cases are associated with \textbf{higher BMI values}, with a higher median and a more concentrated distribution compared to non-stroke cases. Similarly, \textbf{average glucose levels} are notably elevated among stroke cases, exhibiting a right-skewed distribution that suggests hyperglycemia as a potential risk factor.

Age analysis indicates a sharp increase in stroke prevalence after 60 years, reinforcing its role as a critical risk factor. Additionally, \textbf{hypertension, heart disease, and smoking status} show strong associations with stroke occurrence, with individuals in these categories experiencing higher stroke rates.

While \textbf{gender differences} in stroke prevalence appear minimal, variations are observed across occupational groups, with private-sector employees and self-employed individuals showing slightly higher stroke rates.

These findings provide a foundation for feature selection and model development in stroke prediction using machine learning techniques.

\section{Machine Learning Models and Optimization}

In this study, we employed a comprehensive approach to model selection and hyperparameter tuning to identify the most effective machine learning models for predicting stroke occurrences.

\subsection{Model Selection}
We considered five machine learning models: Random Forest Classifier, Support Vector Machine (SVM), Logistic Regression, Decision Tree Classifier, and Extreme Gradient Boosting (XGB) Classifier \cite{9}. Each model was chosen for its unique strengths and ability to handle different types of data and relationships.

\subsection{Hyperparameter Tuning}
To enhance model performance, we employed hyperparameter tuning using RandomizedSearchCV. This method efficiently explores the hyperparameter space by randomly sampling a subset of possible configurations, reducing computational cost while maintaining effectiveness. The search space was carefully defined for each model, including key hyperparameters such as the number of estimators and depth for tree-based models, kernel type and regularization strength for SVM, and penalty terms for Logistic Regression. The optimal hyperparameters were determined based on cross-validation performance, ensuring that each model achieved the best possible generalization to unseen data (Table \ref{tab:hyp}).

\renewcommand{\arraystretch}{1.2} 
\setlength{\tabcolsep}{8pt} 
\definecolor{lightgray}{gray}{0.9} 

\begin{table*}[ht]
    \centering
    \caption{Best Hyperparameters for Different Models Under Various Sampling Strategies}
    \vspace{5pt} 
    \resizebox{0.8\textwidth}{!}{ 
    \rowcolors{2}{lightgray}{white} 
    \begin{tabular}{|l!{\vrule width 1.5pt}l|l|l|} 
        \specialrule{1.5pt}{0pt}{0pt} 
        \rowcolor{gray!30} 
        \textbf{Model} & \textbf{Undersampling Parameters} & \textbf{SMOTE Parameters} & \textbf{Oversampling Parameters} \\
        \specialrule{1.5pt}{0pt}{0pt} 
        Random Forest & \begin{tabular}[c]{@{}l@{}}n\_estimators=10,\\max\_depth=20,\\min\_samples\_split=5,\\min\_samples\_leaf=2\end{tabular} & \begin{tabular}[c]{@{}l@{}}n\_estimators=100,\\max\_depth=None,\\min\_samples\_split=2,\\min\_samples\_leaf=1\end{tabular} & \begin{tabular}[c]{@{}l@{}}n\_estimators=100,\\max\_depth=None,\\min\_samples\_split=2,\\min\_samples\_leaf=1\end{tabular} \\
        \hline
        SVM & \begin{tabular}[c]{@{}l@{}}kernel='linear',\\gamma=1,\\degree=3, \\C=0.1\end{tabular} & \begin{tabular}[c]{@{}l@{}}kernel='rbf',\\gamma=10,\\degree=3, \\C=10\end{tabular} & \begin{tabular}[c]{@{}l@{}}kernel='rbf',\\gamma=10,\\degree=3, \\C=10\end{tabular} \\
        \hline
        Logistic Regression & \begin{tabular}[c]{@{}l@{}}solver='lbfgs',\\penalty='l2',\\C=0.1\end{tabular} & \begin{tabular}[c]{@{}l@{}}solver='lbfgs',\\penalty='l2',\\C=10\end{tabular} & \begin{tabular}[c]{@{}l@{}}solver='lbfgs',\\penalty='l2',\\C=1\end{tabular} \\
        \hline
        Decision Tree & \begin{tabular}[c]{@{}l@{}}max\_depth=5,\\min\_samples\_split=2,\\min\_samples\_leaf=1,\\criterion='entropy'\end{tabular} & \begin{tabular}[c]{@{}l@{}}max\_depth=10,\\min\_samples\_split=10,\\min\_samples\_leaf=4,\\criterion='entropy'\end{tabular} & \begin{tabular}[c]{@{}l@{}}max\_depth=10,\\min\_samples\_split=10,\\min\_samples\_leaf=4,\\criterion='entropy'\end{tabular} \\
        \hline
        XGBoost & \begin{tabular}[c]{@{}l@{}}n\_estimators=100,\\max\_depth=7,\\learning\_rate=0.01,\\subsample=0.8\end{tabular} & \begin{tabular}[c]{@{}l@{}}n\_estimators=100,\\max\_depth=3,\\learning\_rate=0.3,\\subsample=0.8\end{tabular} & \begin{tabular}[c]{@{}l@{}}n\_estimators=100,\\max\_depth=3,\\learning\_rate=0.3,\\subsample=0.8\end{tabular} \\
        \specialrule{1.5pt}{0pt}{0pt} 
    \end{tabular}
    }
\label{tab:hyp}
\end{table*}

\section{Results and Discussion}

This section presents the performance of different machine learning models using various balancing techniques. The results are analyzed in terms of accuracy, precision, recall, F1-score, and cross-validation performance.

\subsection{Model Performance Evaluation}
The models were evaluated using three different data balancing techniques:{Oversampling, SMOTE (Synthetic Minority Over-sampling Technique), and Undersampling}. The results indicate variations in performance across different techniques and models.

\begin{itemize}
    \item \textbf{Random Forest (RF)} and \textbf{Support Vector Machine (SVM)} exhibited the highest accuracy across different balancing techniques, particularly with \textbf{Oversampling and SMOTE}.
    \item \textbf{Logistic Regression (LogReg)} showed lower accuracy compared to tree-based models but remained competitive in balanced data scenarios.
    \item \textbf{Decision Tree (DT)} and \textbf{XGBoost (XGB)} provided stable results, with XGBoost outperforming Decision Tree in most cases.
\end{itemize}

\subsection{Impact of Data Balancing Techniques}
Each balancing technique affected the models' performance differently:

\begin{itemize}
    \item \textbf{Oversampling:} Models achieved the highest test accuracy, with \textbf{SVM (99.28\%)} and \textbf{Random Forest (99.02\%)} performing the best. XGBoost also performed well, achieving \textbf{91.87\% test accuracy}.
    \item \textbf{SMOTE:} Test accuracies slightly dropped compared to oversampling, with \textbf{Random Forest (93.98\%)} and \textbf{SVM (93.52\%)} still maintaining strong performances. XGBoost improved compared to Decision Tree, showing an accuracy of \textbf{93.06\%}.
    \item \textbf{Undersampling:} All models showed a significant drop in performance. \textbf{Random Forest’s accuracy decreased to 74.00\%}, and \textbf{XGBoost dropped to 83.00\%}. Logistic Regression and SVM were more resilient, with results around \textbf{74-75\% accuracy}.
\end{itemize}

\begin{figure}[b]
\centering
\includegraphics[width=\columnwidth]{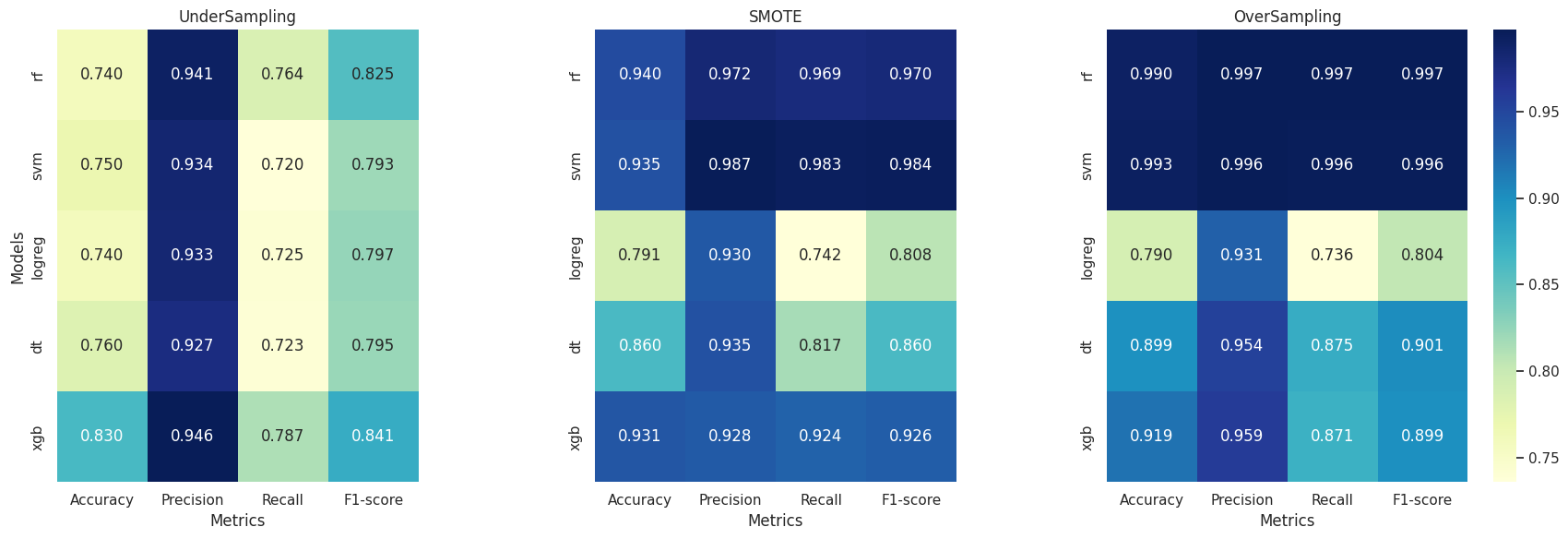}
\caption{Summary results from models}
\label{fig:sum_res}
\end{figure}

These results (Fig. \ref{fig:sum_res}) highlight that {Oversampling and SMOTE are more effective in improving model performance, while Undersampling may not be suitable for datasets with severe class imbalances}.

\subsection{Comparative Analysis of Classification Reports}
From the classification reports, we analyze the models in terms of precision, recall, and F1-score:


\begin{itemize}
    \item \textbf{Random Forest (Oversampling):} Achieves excellent results across all metrics, with Precision, Recall, and F1-score all at \textbf{0.997}.

    \item \textbf{Logistic Regression} underperforms consistently across all techniques. For example, in Oversampling it achieves \textbf{F1-score = 0.804}, \textbf{0.808} in SMOTE and \textbf{0.797} in Undersampling.

    \item \textbf{Decision Tree} delivers moderate results. In Oversampling, it achieves \textbf{Precision = 0.954}, \textbf{Recall = 0.875}, and \textbf{F1-score = 0.901}, which are lower than Random Forest and XGBoost.

    \item \textbf{XGBoost} shows strong recall across all techniques. In SMOTE it achieves \textbf{Precision = 0.928}, \textbf{Recall = 0.924}, and \textbf{F1-score = 0.926}, making it suitable when sensitivity to positive cases is a priority.

\end{itemize}

\subsection{Feature Importance Analysis}
\begin{figure}[b]
\centering
\includegraphics[width=\columnwidth]{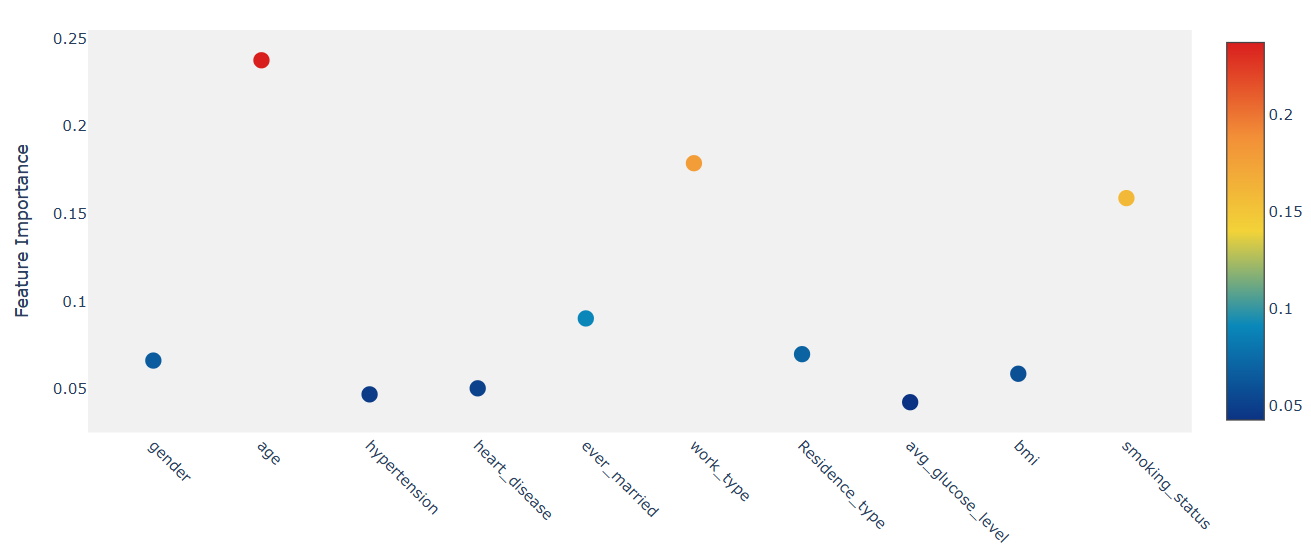}
\caption{XGBoost Model Feature Importance}
\label{fig:XGB_features}
\end{figure}

Feature importance analysis (Fig. \ref{fig:XGB_features}) was conducted for tree-based models (Random Forest, Decision Tree, and XGBoost) to identify the most influential features:

\begin{itemize}
    \item {Age consistently showed the highest importance across all models}.
    \item {Average Glucose Level and BMI were also highly important}, indicating their strong contribution to classification decisions.
    \item {Less significant were Heart Disease and Ever Married features}.
\end{itemize}

This analysis provides insights into which features are most relevant for classification tasks and can be used for feature selection in future studies \cite{6} \cite{9}.

\subsection{Discussion on Model Selection and Generalization}
The results indicate that {ensemble models (Random Forest and XGBoost) are more effective in handling class imbalances}, achieving high accuracy and stable cross-validation scores. SVM also performs exceptionally well, particularly in oversampling and SMOTE scenarios. {Logistic Regression struggles to maintain high performance, which suggests that linear models may not be well suited for this dataset.}

Additionally, {XGBoost offers a balanced trade-off between accuracy and interpretability, making it a robust model choice for real-world deployment}.

\section{Feature Importance Analysis for Elderly Patients (65-80 Years)}

Restricting the dataset to patients aged 65 to 80 years altered the significance of certain features. Across all models, work type (Feature 7) and glucose level (Feature 8) became more influential, while other health-related factors gained prominence compared to the full dataset.

In Random Forest, work type and glucose level were the strongest predictors. Decision Tree showed increased importance for smoking status (Feature 9), suggesting its relevance in elderly patients. XGBoost highlighted hypertension (Feature 2) and heart disease (Feature 3) as more significant than in the full dataset, aligning with expectations that cardiovascular conditions play a crucial role in this age group.

\begin{figure}[htbp]
\centering
\includegraphics[width=\columnwidth]{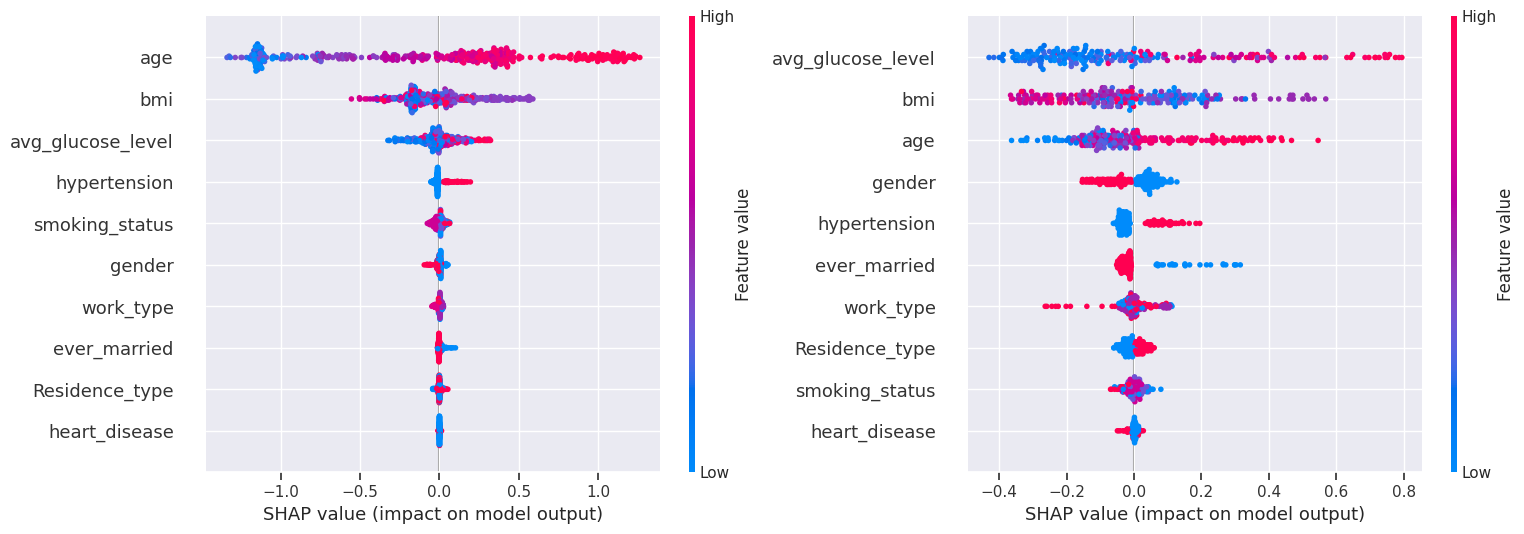}
\caption{SHAP summary plots for the XGBoost model}
\label{fig:SHAP}
\end{figure}

SHAP analysis of the XGBoost model (Fig. \ref{fig:SHAP}) further supports these findings. In the full dataset, age, BMI, and glucose level dominate stroke risk prediction. When restricted to elderly patients, glucose level and BMI gain importance, while hypertension and heart disease become stronger predictors, reflecting their critical role in stroke risk. These results reinforce the need for age-specific predictive models to enhance clinical applicability.

\section{Conclusion}



This study underscores the potential of machine learning in stroke risk prediction, addressing key challenges such as class imbalance and feature selection. Our findings highlight the significant role of demographic, clinical, and lifestyle factors, with age, average glucose level, and BMI emerging as key predictors. Notably, the analysis reveals the importance of age-specific models, as the predictive influence of factors shifts across different age groups. In elderly patients (65–80 years), work type and glucose level become more influential, while hypertension and heart disease gain prominence.

By achieving high predictive accuracy and identifying impactful features, this research supports the development of stroke risk assessment tools with potential integration into clinical decision systems. These tools could assist clinicians in early intervention planning and personalized prevention strategies. However, challenges such as data variability, model interpretability, and deployment in real-world healthcare settings remain. Future research should focus on improving model sensitivity, incorporating diverse datasets, and validating predictions in clinical environments. Advancing these models could enhance early detection strategies, ultimately improving patient outcomes and stroke prevention.

\vspace{12pt}

\end{document}